\documentclass[]{fairmeta}
\pdfoutput=1 
\usepackage{xspace}

\usepackage{amssymb}
\usepackage{pifont}

\usepackage{graphicx}
\usepackage{subcaption}
\usepackage{float}
\usepackage{pythonhighlight}
\usepackage{enumitem}
\usepackage[utf8]{inputenc}

\newcommand{\BTSlong}{\textsc{Branch-Train-Stitch}\xspace}

\newcommand{\BTS}{BTS\xspace}  

\title{\BTS: Harmonizing Specialized Experts into a Generalist LLM}

\author[1,2,*]{Qizhen Zhang} 
\author[1,\diamond]{Prajjwal Bhargava}
\author[1,\diamond]{Chloe Bi}
\author[1,\diamond]{Chris X. Cai}
\author[1,2,\diamond]{Jakob Foerster}
\author[1,\diamond]{Jeremy Fu}
\author[1,\diamond]{Punit Singh Koura}
\author[1,\diamond]{Ruan Silva}
\author[1,\diamond]{Sheng Shen}
\author[1,\dagger]{Emily Dinan}
\author[1,\dagger]{Suchin Gururangan}
\author[1,\dagger]{Mike Lewis}

\affiliation[1]{GenAI at Meta}
\affiliation[2]{University of Oxford}

\contribution[*]{First author, work done at Meta}
\contribution[\diamond]{Ordered alphabetically}
\contribution[\dagger]{Joint last author}

\abstract{
We present \BTSlong (BTS), an efficient and flexible training algorithm for combining independently trained large language model (LLM) experts into a single, capable generalist model. Following \cite{li2022branch}, we start with a single ``seed'' language model which is branched into domain-specific (e.g., coding or math) experts with continual pretraining. \BTS combines experts into a generalist model using lightweight stitch layers, which are inserted between frozen experts and the seed LLM, and trained on a small datamix of the expert domains. Stitch layers enable the seed LLM to integrate representations from any number of experts during the forward pass, allowing it to generalize to new domains, despite remaining frozen. Because BTS does not alter the constituent LLMs, \BTS provides a modular and flexible approach: experts can be easily removed and new experts can be added with only a small amount of training. Compared to alternative model merging approaches, \BTS yields the best generalist performance on a variety of downstream tasks, retaining the specialized capabilities of each of the experts.  
}

\date{\today}
\correspondence{Qizhen Zhang at \email{qizhen.zhang@eng.ox.ac.uk}}

\begin{document}

\maketitle

\section{Introduction}
\begin{figure}[!t]
  \centering
\includegraphics[width=0.85\columnwidth]{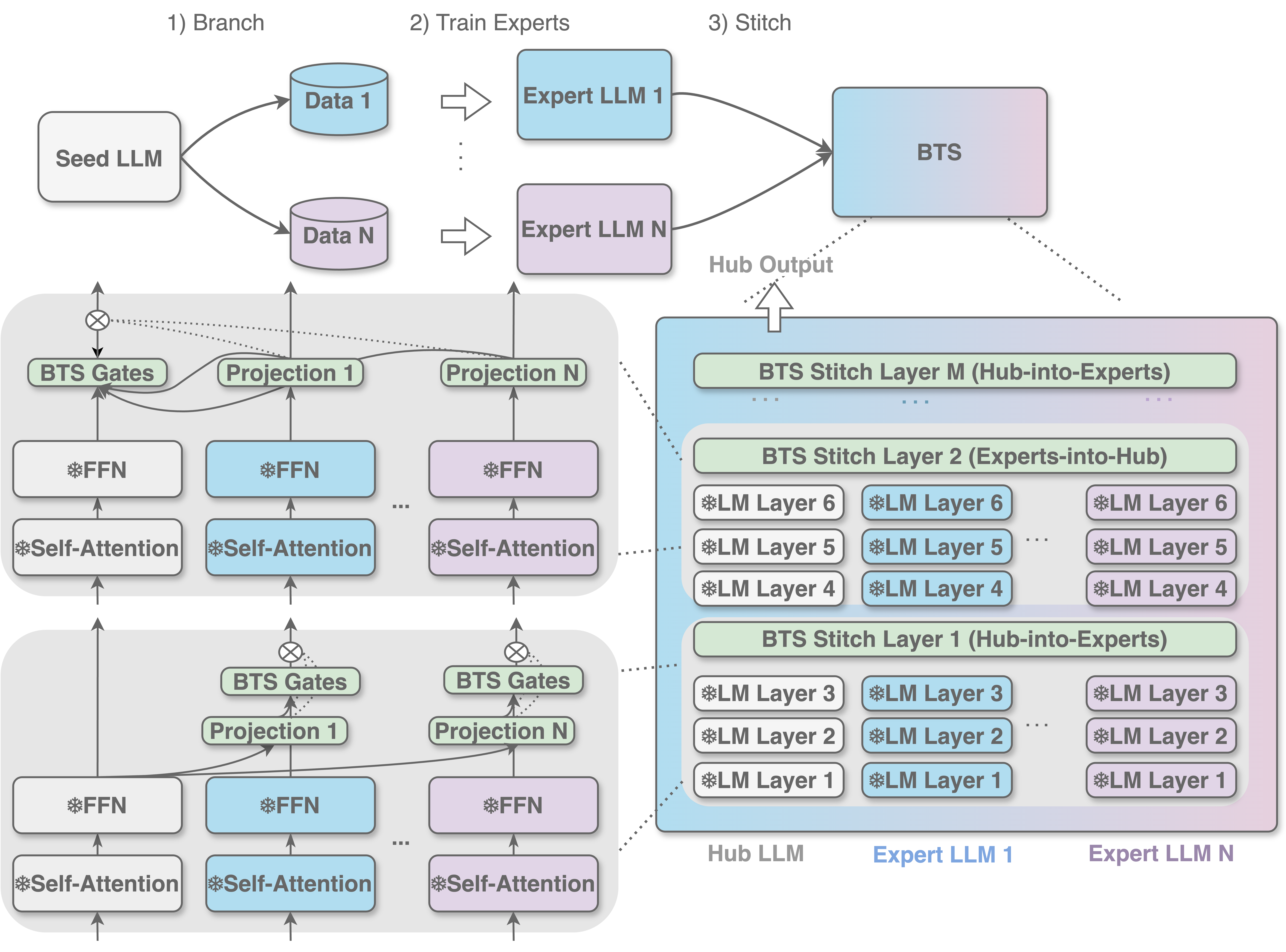}
  \caption{\textbf{Overview of the \BTS algorithm.}
\BTS operates in three phases. Different colors correspond to different expert domains.
\textbf{1) Branch}: Following \citet{li2022branch}, we begin with a pretrained seed model and create $N$ copies of it. \textbf{2) Train Experts}: Each copy is independently pretrained on its respective data mixture, resulting in specialized expert models, as described in \citet{li2022branch}.
\textbf{2) Stitching}: Stitch layers are inserted throughout the layers, alternating between the \textit{Experts-into-Hub stitch layer} and the \textit{Hub-into-Experts stitch layer}. Only the stitch layers are updated during this training phase.
The \BTS model always have a Experts-into-Hub stitch layer as the last layer, as the hub output is returned as the final \BTS output. 
}
\label{fig:bts}
\end{figure}

To achieve strong performance across diverse domains, large language models (LLMs) are often densely trained on trillions of tokens using thousands of GPUs \citep{dubey2024llama}. Dense training requires substantial resources and significant infrastructure challenges, often requiring massive synchronization across distant compute clusters. Dense training also  poses difficult datamix tradeoffs \citep{sang2023doremi,ye2024data}; 
for example, it can be challenging to improve performance on a new domain without forgetting the original data \citep{mccloskey1989catastrophic,aghajanyan2021betterfinetuning} or debug and correct unwanted behaviors without impacting others \citep{tuan2024towardssafety}.

Expert merging techniques like Branch-Train-Merge \citep[BTM;][]{li2022branch, gururangan2023scaling} address these challenges by asynchronously training distinct expert models, specialized to different domains, and merging them back into a single generalist language model by ensembling them at inference time. Experts can be removed from the mix or added as needed. However, BTM is limited because there are no learned connections between expert layers; this restricts the model's overall expressivity, especially in distant test domains. On the other hand, approaches like  Branch-Train-MiX \citep[BTX;][]{sukhbaatar2024branch}, which upcycles experts into an Mixture-of-Experts (MoE) model \citep{shazeer2017moe}, show strong downstream task performance, but lose the flexibility and interpretability inherent in a modular approach where experts remain distinct and intact.

We present \BTSlong (\BTS), a new algorithm for building a generalist LLM from a collection of smaller expert models which achieves the best generalist model performance. Like other merging techniques \citep{li2022branch, gururangan2023scaling}, \BTS begins with a training phase in which experts are created via independent continued pretraining on domains of interest (starting from a shared ``seed'' checkpoint; \citealt{li2022branch}). After expert training, the experts are adapted into a unified, generalist model by inserting and training \emph{stitch layers} between models, while keeping the experts themselves frozen. 

This stitching architecture adds connections between experts via a gating mechanism on top of the language model layer outputs which determine how hidden states from one expert flow into another. One can imagine several ways to combine representations produced by different experts: all experts can directly connect to all other experts, only certain experts can connect to certain others, and everything in between. We opt for a hub-and-spoke model, in which a central ``hub'' model (the seed LLM) can update its own representations via the spokes (specialized experts), and vice versa, but the experts have no direct connection to each other. This design choice balances efficiency and performance. Since the seed model is trained on a variety of data, it is a natural choice for the hub, so all of our experiments adopt this set-up. 
For each layer in the forward pass, the stitching architecture alternates between \emph{hub-to-expert merging}, where the hidden representations of the experts are updated with a projected hub LLM representation, and \emph{hub-to-expert merging}, where the hub's hidden representation is updated with a combined hidden representation of all experts. The final output provided by the merged LLM is the output of the seed model. These design choices are further motivated and validated empirically with ablations in \autoref{sec:ablations}.

In experiments (\autoref{sec:core_results}), we find that \BTS achieves the best generalist model performance compared to both expert merging and expert upcycling baselines and can even perform better than some individual experts on their target tasks. Notably, this is achieved with training only the small set of stitching parameters. The modular design of \BTS, in which individual experts remain unchanged in the merging process, offers flexibility and interpretability. Targeted performance improvements for specific domains can be achieved completely asynchronously. Furthermore, downstream behaviors can be easily understood by analyzing which experts are `active' at any given token, providing transparency into the model's decision-making process.

Our contributions are summarized as follows:
\begin{itemize}
    \item \textbf{\BTSlong, \autoref{sec:methods}}: We propose \BTSlong, an efficient and flexible approach for stitching distinct expert models into a more powerful, generalist LLM.
    \item \textbf{Experiments, \autoref{sec:core_results}}: We validate this approach through experiments on seed language models of 2.7B parameters. Our results demonstrate that \BTS outperforms competitive baselines in downstream task performance, achieving the best average performance across benchmarks.
    \item \textbf{Analysis, \autoref{sec:ablations}}: We motivate the \BTS architectural choices with ablations and investigate the impact on ``cross capability'' tasks, i.e. tasks at the intersection of expert domains, and show that, in certain settings, \BTS can achieve cross capability performance greater than any expert. Finally, we provide detailed analysis of the behavior of stitch layers at inference time, showing that BTS can dynamically adjust its expert utilization even within the same prompt. 
\end{itemize}

\section{\BTSlong}\label{sec:methods}

This section provides an overview of the \BTS algorithm, beginning with a brief background on language model architectures (\autoref{subsec:background}), followed by a detailed description of the \BTS methodology (\autoref{sec:method_overview}) and architecture (\autoref{sec:model_arch}).

\subsection{Language model architecture background}
\label{subsec:background}
\paragraph{Transformer}
The typical architecture of large language models (LLMs) is built by stacking multiple Transformer blocks \citep{vaswani2017attention}. Each Transformer block consists of a Multi-Headed Attention module, commonly referred to as the \textit{attention layer}, followed by a residual connection and a feed-forward neural network (FFN).

\paragraph{Mixture-of-Experts}
The Mixture of Experts \citep[MoE;][]{shazeer2017moe} model replaces the FFN in the Transformer by an MoE layer. An MoE layer consists of a linear router and a set of $N$ \textit{FFN experts}, denoted as $\left\{\text{FFN}_i(x)\right\}_{i=1}^N$. The router produces normalized router logits $p(x)$ for the input representation $x$, where $p_i(x)$ is the gating value for the $i$-th FFN expert, $\text{FFN}_i$. The router assigns the input representation $x$ to a subset of experts, $\mathcal{T}$, with the highest gating values. The final output of the MoE layer is the weighted sum of the selected experts' outputs, weighted by their gating values:
\begin{equation}
\label{eq:moe}
y_\text{MoE} = \sum_{i \in \mathcal{T}} p_i(x) \text{FFN}_i(x).
\end{equation}

\paragraph{Mixture-of-Attention}
Mixture of Attention \citep[MoA;][]{zhang2022mixture} extends MoE by also replacing the attention layer in Transformers with an MoA layer. Similar to the MoE layer, an MoA layer comprises of a set of $N$ \textit{attention experts} (denoted as $\left\{\text{Attention}_j(x)\right\}_{j=1}^N$), a linear router that outputs normalized router logits $q(x)$. Like the MoE, the MoA layer's final output is a gating-value weighted sum of the computations from the selected attention experts $\mathcal{M}$:
\begin{equation}
\label{eq:moa}
y_\text{MoA}=\sum_{i \in \mathcal{M}} q_i(x) \text{Attention}_i(x) .
\end{equation}

\subsection{\BTS algorithm overview}\label{sec:method_overview}

The \BTS algorithm involves three stages, resulting in an efficiently-trained generalist dense model. The process is visualized in \autoref{fig:bts}. 
\begin{enumerate}
    \item  \textbf{Branch}: Following \citealt{li2022branch}, given a pretrained Transformer seed model $m_0$, we create $n$ copies of the model $m_1, ..., m_n$.
    \item  \textbf{Train}: Also following \citealt{li2022branch}, each copy of the seed model $m_i$ independently undergoes a continued pretraining  phase  on a specialized data mixture, $\mathcal{D}_{i}$, each tailored to different domains such as code, mathematics, and multilingual \citep{gururangan2020dontstoppretrainingadapt}. This phase yields specialized models that have enhanced performance within their respective domains compared to the seed model $m_0$. However, these models might perform worse in domains outside of their specialization as they forget knowledge from the initial pretraining phase. We refer to these models $m_i$ as \emph{experts}, and note that this usage of the term ``expert'' differs in meaning from the FFN / attention experts in MoE and MoA models.
    \item \textbf{Stitch}: We merge the seed ($m_0$) and expert models ($m_i$, $i > 0$) from the previous steps using our lightweight stitch layers $\Psi$, which are trained for a small number of steps on a mixture of data from expert domains. The stitch layer architecture is described in  \autoref{sec:model_arch}. Importantly, \emph{only} the  stitch layers are updated during this phase, while the parameters of the seed and expert models remain frozen. This ensures that \BTS training is a flexible approach --- experts can be added or removed after merging, only requiring retraining stitch layer parameters.
\end{enumerate}

\subsection{Model architecture}
\label{sec:model_arch}

Next, we provide additional details on the BTS architecture (\autoref{fig:bts}). We introduce the \emph{stitch layer}, which, as mentioned above, merges $n + 1$ Transformer models $m_0, ..., m_n$. We designate $m_\text{0}$ as the \emph{hub} and $m_1, ..., m_n$ as the \emph{experts}. The hub is usually the seed model, unless otherwise noted.

Suppose the expert $m_i$ contains $L$ Transformer layers, $\{\ell^j_i\}_{j=1}^{L}$. We insert $K$ stitch layers -- one each after every $ \lfloor \frac{L}{K} \rfloor$ Transformer layers. We denote $\Psi_j$ as the stitch layer inserted after Transformer layers $\{\ell^j_i\}_{i=0}^{n}$. The stitch layer $\Psi_j$, takes as input the hidden states, or outputs, from the hub's $j$-th layer $\ell^j_0$ and the experts' $j$-th layers, $\{\ell^j_i\}_{i=1}^{n}$. We denote the hidden states respectively as $h^{j}_0$ for the hub and $\{h^j_i\}_{i=1}^{n}$ for the experts.
The outputs of the stitch layer, $\Psi_j(h^j_0, \dots, h^j_n) = (\tilde{h}^j_0, \dots, \tilde{h}^j_n)$, become the input to the corresponding experts $m_i$'s $j+1$-th layer ($\ell^{j+1}_i$).

Each stitch layer $\Psi$ introduces two sets of learnable parameters: 
\begin{enumerate}
    \item \textbf{Linear projections}, $\{w_{\text{proj}_1}, ..., w_{\text{proj}_n}\}$, where $w_{\text{proj}_i} \in \mathbb{R}^{\text{dim} \times \text{dim}}$ either projects the expert hidden states to the hub model's hidden state space or projects the hub model’s hidden state into the expert’s hidden state space.
    
    \item \textbf{A linear gate} $w_\text{gate} \in \mathbb{R} ^{\text{dim} \times \text{dim} \times n}$, which computes the contribution of each model's hidden state.
\end{enumerate}
To apply these gates, we alternate between two types of stitch layers (refer to \autoref{fig:bts} for the illustration and \autoref{sec:pseudocode} for the pseudo code):

\paragraph{The Experts-into-Hub Stitch Layer}  In this layer, the expert models' hidden states are first projected into the hub model's hidden state space,. The hub then combines its own representation with the projected experts' hidden states, weighted by the outputs of a softmax-based gating mechanism.

\begin{equation}
    \begin{aligned}
    g &= \operatorname{softmax}(\text{dropout}(w_\text{gate}(h_0))) \\
    \tilde{h}_i &= w_{\text{proj}_i}(h_i) \qquad \qquad \qquad \qquad \qquad \text{for $i \in \{1,..., n\}$} \\
    \tilde{h}_0 &= h_0 *  g_0 + \sum_{i=1}^{n}  g_i * \tilde{h}_i,
    \end{aligned}
\end{equation}
where $g_i$ correspond to the $i$-th expert in the gate value $g$.

\paragraph{The Hub-into-Experts Stitch Layer} 

In this layer, the hub representation is projected into each of the expert model's hidden state space. Each expert combines its own hidden state with a gated projection of the hub representation using a sigmoid-based gating mechanism:

\begin{equation}
    \begin{aligned}
        g &= \operatorname{Sigmoid}(\text{dropout}(w_\text{gate}(h_0))) \\
        \tilde{h}_0 &= h_0 \\
        \tilde{h}_i &= (1-g_{i}) * h_i + g_{i} * w_{\text{proj}_i}(h_0)  \qquad \qquad \text{for $i \in \{1,..., n\}$} \\
    \end{aligned}
\end{equation}

As we demonstrate in \autoref{sec:ablations}, this alternating architecture is essential for enabling cross capabilities without degrading generalist performance.

\section{Results: Building a generalist model}\label{sec:core_results}
We validate the \BTS approach through experiments with a seed language models of 2.7B parameters. We describe model (\autoref{sec:model_setup}), data (\autoref{sec:data_details}), baseline (\autoref{subsec:baselines}), and evaluation (\autoref{subsec:evaluation}) details and discuss experimental results in \autoref{sec:results}.

\subsection{Model details}
\label{sec:model_setup}

\paragraph{Seed model}
We pretrain a 2.7B parameter language model, following the same text recipe used in Llama 3 \citep{dubey2024llama}. See \autoref{tab:model_configs} for architecture details.
We employ a learning rate schedule that warms up from $0$ to 4e-4 over 2000 steps, then undergoes a cosine decay to 1\% of the peak learning rate. The seed model is trained for 2.2M steps on 15T tokens.

\begin{table}[h!]
\centering
\renewcommand{\arraystretch}{1.3}
\begin{tabular}{l | c}
\toprule
\textbf{Layers} & 20 \\
\textbf{Model Dimension} & 3072 \\
\textbf{FFN Dimension} &  12288\\
\textbf{Attention Heads} & 24\\
\textbf{Key/Value Heads} & 1 \\
\textbf{Activation Function} & {SwiGLU} \\
\textbf{Vocabulary Size} & {128,000} \\
\textbf{Positional Embeddings} & {{RoPE} ($\theta = 500,000$)} \\
\bottomrule
\end{tabular}
\caption{\textbf{Architecture details} for the 2.7B parameter seed model and expert models.}
\label{tab:model_configs}
\end{table}

\paragraph{Expert models}
We create three copies of the seed model, each of which is continually trained for 96k training steps over a 200B token specialized data mixture to produce expert models for code, mathematics, and multilingual tasks. During the continued pretraining phase, we use a batch size of 2M tokens and a learning rate of 5e-6, followed immediately by a cosine decay schedule that reduces the learning rate to 1\% of its initial value. This learning rate is derived by annealing from the final learning rate used at the end of seed model pretraining, adjusted to account for the reduced batch size in this continued pretraining phase. We adopted this learning rate strategy as it yielded the most stable learning during the continued pretraining phase

\paragraph{\BTS model}
We use four stitch layers to combine the seed model together with the three expert models. The four stitch layers are inserted after every five layers in the seed and expert models. We refer to the resulting model as the \BTS model. As described in \autoref{sec:methods}, the four stitch layers alternate between a \textit{Merge-into-Expert layer} and \textit{Experts-into-Hub stitch layer}.  Upon initialization, the \BTS model is further trained for 15B tokens over 7000 steps using a batch size of 2M tokens. The optimization objective is to minimize the next-token prediction loss from the hub model's output. The learning rate schedule warms up from $0$ to 5e-6 over 2000 steps, then undergoes a cosine decay to 1\% of the peak learning rate. Note that during the \BTS training phase, only the stitch layers are updated while all the parameters of the seed model and the expert models are frozen. 

\subsection{Data details}
\label{sec:data_details}
\paragraph{Seed model} We adopt the same text pretraining mixture as Llama 3 \citep{dubey2024llama}.

\paragraph{Expert models}
In the \textbf{continued pretraining phase}, each dense expert is trained on a specialized data mixture for 200B tokens:
\begin{itemize}
  \item \textbf{Code}: We adopt a recipe similar to that of CodeLlama \citep{rozier2023code} with $>85\%$ code tokens, utilizing the code data subset of the seed model mixture.
  \item \textbf{Math}: We continue pretraining on the OpenWebMath dataset \citep{paster2023openwebmath}. 
  \item \textbf{Multilingual}: We utilize a mixture of $90\%$ non-English data and $10\%$ English data, with each subset pulled from the seed model mixture, following the multilingual expert recipe described in \citet{dubey2024llama}.
\end{itemize}

\paragraph{\BTS model}
The data mixture for the \BTS training phrase consists of $15\%$ expert domain tokens for each of the code, math, and multilingual domains. The remaining $55\%$ of the mixture consists of the pretraining data utilized for the seed model outside of these domains.

\begin{table}[t!]
\centering
\renewcommand{\arraystretch}{1.3}
\begin{tabular}{l  c  c  c }
\toprule
& \textbf{Training Params} & \textbf{Total Params} & \textbf{Active 
Params}\\
\midrule
\multicolumn{3}{l}{\emph{Expert upcycling}} \\
BTX Sample & 7.2B & 7.2B & 2.9B \\
BTX Soft & 7.2B & 7.2B & 7.2B\\
BAM & 8.4B & 8.4B & 8.4B \\
\midrule 
\multicolumn{3}{l}{\emph{Expert merging}} \\ 
Model Soup & N/A & 2.7B & 2.7B\\
BTM & N/A & 10.8B & 10.8B \\
Expert Routing & 15k & 10.8B & 2.7B\\
BAM Adapters & 1.5B & 9.9B & 9.9B \\
BTS & 264M & 11B & 11B \\
\bottomrule
\end{tabular}
\caption{\textbf{Training, total, and active parameter count} for \BTS and baselines. We use \textit{``expert upcycling''} to describe MoE upcycling methods where the seed and experts themselves do not remain intact during the MoE training phase. These methods require significantly more training parameters, and thus are less modular, less flexible, and less interpretable. We use \textit{``expert merging''} to describe methods, such as \BTS, where the seed and expert models remain frozen during the merging phase. Expert merging methods require minimal number of training parameters, making them more modular and interpretable.}
\label{tab:param_count}
\end{table}

\subsection{Baselines}\label{subsec:baselines}

In addition to the seed and expert models, we also compare \BTS with \textit{ expert upcycling} and \textit{expert merging} baselines. We use \textit{expert upcycling} to describe methods where the seed and expert models are used to initialize an MoE model, which is further trained. The entire MoE is updated during training and as such the experts and seed model themselves do not remain intact. This approach loses the flexibility and interpretability inherent in a more modular approach, and any model change requires updating a large number of parameters.
On the other hand, we use \textit{expert merging} to describe methods, such as \BTS, in which the seed and expert parameters remain frozen during the merging phase.

\paragraph{\textbf{Expert upcycling baselines:}}
\begin{itemize}
    \item \textbf{BTX \citep{sukhbaatar2024branch}:} We upcycle the seed model and three expert models into an MoE. Our baselines include two BTX variants, where the FFN experts employ one of two routing strategies: 1) \textit{sample top-1 routing} \citep{sukhbaatar2024branch}, where we use a Gumbel-Softmax \citep{jang2016categorical} for the routing function, and 2) \textit{soft-routing}, where all four experts are activated at all times. We use the same experimental setup as \BTS runs, including training data and the learning rate schedule. See \autoref{subsec:background} for details on the MoE architecture.

    \item \textbf{BAM \citep{zhang2024bam}:} We upcycle the seed model and the three expert models into an MoE with both attention experts and FFN experts. See \autoref{subsec:background} for a description of the attention experts architecture. We employ soft-routing for both sets of experts, ensuring that, like \BTS, all FFN and attention parameters of the seed and expert models are activated during training and inference.  We use the same experiment setup as \BTS runs.
\end{itemize}

\paragraph{\textbf{Expert merging baselines:}}
\begin{itemize}
    \item \textbf{Model soup \citep{wortsman2022model}:} We uniformly average the weights of the seed and expert models. Unlike other baselines, no further training is required upon initialization.

    \item \textbf{BTM \citep{li2022branch}:} We ensemble the output logits of the seed and expert models. The ensemble weights are estimated using Bayes’ rule with a uniform prior \citep{li2022branch,gururangan2023scaling}. Like the model soup baseline, no further training is required upon initialization.
    
    \item \textbf{Expert routing:} 
    We train a linear router $\in \mathbb{R}^{\text{dim} \times n}$ that routes to either the seed model or one of the expert models. The router's training objective is a classification cross-entropy loss where the target is the model with the smallest next-token prediction loss for the input. Given a prompt, the router decides on the model and routes all subsequent tokens to the same model. During training, the routing decision is made based on the average embedding of the first $t$ tokens in the input, where $t$ is randomly sampled between 32 and 256. During inference, the routing decision is made based on the average embedding of the entire prompt. We train the linear router with a constant learning rate of 5e-4 and batch size of 1M. The model is trained for 1B tokens only, as we did not see an improvement in downstream metrics or training loss with further training. 
    \item \textbf{BAM with adapters \citep{zhang2024bam}:} We train an expert-intact variant of BAM with soft-routing, which we refer to as BAM with adapters. In this variant, each attention expert and each FFN expert's output undergo a linear adapter layer
    $W_{\text{proj}_i} \in \mathbb{R}^{dim \times dim}$. Formally, we replace \autoref{eq:moe} and \autoref{eq:moa} by the following:
    \begin{equation}
    \begin{aligned}
    y_\text{MoE} &= \sum_{i \in \mathcal{T}} p_i(x) W_{\text{ffn proj}_i} \left(\text{FFN}_i(x)\right) \\
    y_\text{MoA} &=\sum_{i \in \mathcal{M}} q_i(x) W_{\text{attn proj}_i} \left( \text{Attention}_i(x) \right) .
    \end{aligned}
    \end{equation}
    Only the router and adapters are updated during training, while all other parameters remain frozen. We use the same experiment setup as BTS runs.
\end{itemize}

We show a comparison of the number of training, active, and total parameters in \autoref{tab:param_count}. Note that \BTS has the most total parameters of all variants, but only a small fraction of the training parameters of the expert upcycling variants.

\subsection{Evaluation}\label{subsec:evaluation}
We assess model performance with zero-shot and few-shot downstream tasks relevant to the expert domains.

\begin{itemize}
    \item \textbf {General Knowledge and Reasoning}: To assess general knowledge and reasoning capabilities, we report MMLU \citep[5-shot;][]{hendrycks2021mmlu} and Big-Bench Hard \citep[3-shot;][]{suzgun2022challenging}. In tables, we denote Big-Bench Hard as BBH.
    \item \textbf{Code}: For code generation capabilities, we evaluate on MBPP \citep[3-shot;][]{austin2021program} and HumanEval \citep[0-shot;][]{chen2021codex} benchmarks. We denote  HumanEval as HE in the results table for brevity.
    \item  \textbf{Multilingual}: For measuring multilingual capabilities, we use machine translation sub-tasks in Flores \citep[1-shot;][]{goyal2022flores}. Specifically, we evaluate on seven languages:
    Dutch, Spanish, Portuguese, Vietnamese, Indonesian, Hindi, and French. We display the sub-tasks evaluations into two categories, 1) those with English as the source translation language (S), and 2) those with English as the target translation language (T).
    \item \textbf{Math}: For mathematical reasoning, we report the performance on GSM8K \citep[8-shot;][]{cobbe2021training} and MATH \citep[4-shot;][]{hendrycksmath2021}.
\end{itemize}

\subsection{Results}\label{sec:results}

\begin{table}[t!]
\centering
\begin{tabular}{lrrrrrrrrrr}
\toprule
& \multicolumn{2}{c}{\textbf{General}} & \multicolumn{2}{c}{\textbf{Code}} & \multicolumn{2}{c}{\textbf{Multilingual}} & \multicolumn{2}{c}{\textbf{Math}} & \\
\cmidrule(lr){2-3} \cmidrule(lr){4-5} \cmidrule(lr){6-7} \cmidrule(lr){8-9}
 & \textbf{MMLU} &
 \textbf{BBH} & \textbf{MBPP} & \textbf{HE} & \textbf{Flores(S)} & \textbf{Flores(T)} & \textbf{GSM8K} & \textbf{MATH} & \textbf{Avg.} \\ 
\midrule
\emph{\textbf{2.7B Dense models}} \\ 
Seed Model & 28.4 & 35.6 & 27.0 & 20.7 & 29.5 & 35.7 & 10.5 & 4.82 & 24.0 \\
Code Expert & 30.3 & 35.2 & \textbf{32.0} & $^*$\textbf{25.0} & 29.0 & 35.5 & 11.4 & 4.40 & 25.4\\
Multiling. Expert & 26.6 & 34.7 & 26.2 & 18.3 & $^*$\textbf{31.9} & $^*$\textbf{37.1} & 10.8 & 4.16 & 23.7\\
Math Expert & $^*$\textbf{36.3} & $^*$\textbf{37.2} & 26.2 & 16.5 & 23.6 & 32.7 & $^*$\textbf{20.5} & \textbf{10.1} & \textbf{25.4} \\
\midrule 
\emph{\textbf{Expert upcycling}} \\ 
BTX Sample & 30.4 & 36.6 & 30.0 & 21.3 & 30.5 & 36.0 & 13.9 & 6.58 & 25.7\\ 
BTX Soft &34.7&36.8&29.6&23.2& 31.0 & 36.0 & 19.2 & 9.10 & 27.4\\
BAM & 35.2& \textbf{37.1}& 29.8& 22.6& 31.0	& 36.1& \textbf{20.3} & 10.1 & 27.8\\
\midrule
\emph{\textbf{Expert merging}} \\ 
Model Soup & 30.7 & 37.0 & 29.6 & 22.6 & 29.5 & 36.2 & 13.6 & 6.46 & 25.7\\ 
BTM & 30.6 & 37.0 & 31.8 & \textbf{23.8 }& \textbf{31.8} &\textbf{ 37.0} & 12.7 & 10.1 & 26.9\\
Expert Routing & 28.4 & 35.6 & 27.0 & \textbf{23.8} & 30.8 & \textbf{37.0} & 10.5 & 5.04 & 24.8\\ 
BAM Adapters & 34.0 & 37.0 & 28.8& 22.6 & 31.0 & 36.1 & 18.8 & 10.0 & 27.3\\
\BTS &  \textbf{35.8} & 36.9 & $^*$\textbf{32.2} & 22.0 & 30.9 & 36.2 & 20.2 & $^*$\textbf{10.6} & $^*$\textbf{28.1}\\
\bottomrule
\end{tabular}
\caption{\textbf{Performance of \BTS against expert merging and upcycling methods, seed and expert models} measured on popular benchmarks across several capabilities. \textbf{Bolded} numbers indicate the best performance among dense models or merged models, while an asterisk ($^*$) denotes the best performance across all models. See \autoref{subsec:evaluation} for benchmark details. Although dense expert models sometimes achieve the best results in their specialized domains, they often significantly under-perform in other domains. Among all merged models, \BTS achieves the best average performance. Notably, \BTS not only emerges as the most well-rounded generalist expert but also outperforms the corresponding domain-specific experts on MATH and MBPP tasks.}
\label{table:main_results}
\end{table}

Results on general knowledge, code, multilingual, and math benchmarks for the seed model, expert models, and all expert merging and expert upcycling baselines are reported in \autoref{table:main_results}. We make the following observations:
\begin{itemize}
    \item \textbf{Expert models highlight datamix tradeoffs:} While the dense expert models typically achieve the best results in their respective target domains, they often significantly underperform in other domains, highlighting that improving performance in one domain may come at the cost of regressing in others. For example, the Math expert outperforms all models in GSM8K, but lags behind the seed model substantially in coding tasks. 
    \item \textbf{Learned connections are important for expressive merging:} Methods like BAM with adapters and \BTS outperform expert merging methods without learned connections between experts, such Model Soup, BTM, and Expert Routing. This demonstrates the importance of adding learned, intermediate connections between experts.
    \item \textbf{\BTS achieves the best generalist performance:} Among all model variants -- seed, expert, expert merging, and experts upcycling -- \BTS achieves the best average performance across tasks. Notably, \BTS achieves similar or better performance to the expert upcycling baselines at only a fraction of the training parameters.
    \item \textbf{\BTS can outperform individual experts in their specialized tasks:} \BTS emerges not only as the most well-rounded generalist model, but is also the \textit{only} model which achieves \textit{better} performance than any individual expert in some tasks. \BTS outperforms the Code expert in MBPP and the Math expert in the MATH task.  
\end{itemize}

\section{Ablations and analysis}
\label{sec:ablations}

We provide detailed ablations and analysis as follows:
\begin{itemize}
    \item \textbf{Enabling cross capabilities}, \autoref{sec:results_cross_capabilities}: We evaluate how \BTS performs on cross capabilities, or capabilities at the intersection of two or more expert specialties and compare to other merging techniques.
    \item \textbf{\BTS architecture design}, \autoref{sec:arch_design}: We empirically validate several \BTS architecture choices, including assessing the impact of the number of stitch layers, the alternating stitch layer design, and choice of hub model.
    \item \textbf{Interpreting the \BTS stitch layers}, \autoref{sec:stitch_interpretation}: Finally, we provide visualizations and analysis of how the \BTS stitch layer gate values behave at inference time for various downstream tasks.
\end{itemize}

\subsection{Enabling cross capabilities}\label{sec:results_cross_capabilities}

In addition to evaluating merged models on the union of the expert capabilities, we also explore whether merged models can demonstrate entirely new capabilities at the \textit{intersection} of expert specialties \citep{zhong2024law}. For example -- can a Russian-language expert and a Math expert be combined in such a way that the merged model performs better than either expert at Russian math tasks? We refer to these as cross capabilities.

\subsubsection{Cross capabilities experimental set-up} 

\begin{table}[t]
\centering
\begin{tabular}{lrrr>{\columncolor[gray]{0.9}}r}
\toprule
& & \multicolumn{2}{c}{\textbf{Flores}} \\
& \textbf{GSM8K} & \textbf{En/Ru} & \textbf{Ru/En} &  \textbf{Ru-MGSM} \\ 
\midrule
\emph{\textbf{Dense models}} \\ 
Seed Model & 10.5	& 22.8& 32.8& 12.8\\
Math Expert & $^*$\textbf{20.5}	& 10.2& 28.9& 10.8\\ 
Russian Expert &  9.48 & $^*$\textbf{32.3} &  \textbf{34.6}	& 9.60\\
Seed Model (DM) & 12.6	& 24.8	& 32.8	& \textbf{14.0}\\
\midrule 
\multicolumn{3}{l}{\textbf{Expert upcycling}} \\ 
BTX Sample & 15.6 & 29.9	& 34.3 & 17.6\\ 
BTX Soft & 17.6	&  30.6	&  34.5	& 17.6 \\
BAM & \textbf{19.3}	& 30.9	& 34.5	&  $^*$\textbf{18.4} \\
\midrule
\multicolumn{3}{l}{\textbf{Expert merging}} \\ 
Model Soup & 17.5	&14.7	& 32.3	& 13.2\\
BTM & 20.5 & $^*$\textbf{32.3}	& 34.6	& 9.60 \\
Expert Routing & 9.48 & $^*$\textbf{32.3}	& 34.6	& 9.60 \\ 
BAM Adapters & 15.2 &	31.0 &	34.3 &	15.6\\
\BTS  & 13.3 & 31.9 &$^*$\textbf{34.7}	& 16.0 \\
\bottomrule
\end{tabular}
\caption{\textbf{Cross capability performance.} We evaluate the seed model, Russian-language, and Math experts on the Russian subset of MGSM \citep{shi2022language}. We compare their performance with expert merging and expert upcyling baselines trained with small amounts of in-domain data on Russian mathematics. We also continued pretraining the strongest dense model, the seed model on the same in domain data. We call the resulting baseline  ``Seed Model Data Matched (DM)''. \textbf{Bolded} numbers indicate the best performance among dense models or merged models, while an asterisk ($^*$) denotes the best performance across all models. \BTS outperforms the data-matched seed model, and achieves the best cross capability performance among all expert merging methods. This demonstrates that with only a small amount of in-domain data, \BTS models can effectively learn how to combine expert capabilities.}
\label{table:cross_capabilities}
\end{table}

In order to evaluate cross capabilities, we train an additional Russian-language expert specifically on Russian data, and all merged models are created with \emph{only} the Russian and Math experts.  We make these choices in order to study cross capability emergence in a controlled setting: 
\begin{itemize}
\item \textbf{Reducing cross capability expert contamination:} We found that our coding data contained significant portions of non-English natural language, affecting the Code expert's ability in multilingual reasoning tasks, so we remove this model from this mix \citep{blevins2022language}. We further remove the seed model which contains both multilingual and math data. 
\item \textbf{Prevalance of cross capability training and evaluation data:} We limit our study to languages in which we have cross capability data to both train and evaluate the models on --- for this reason, we focused on Russian and Math. 
\end{itemize}

Note that when merging only two experts, there is no notion of ``hub'' model: the stitch layers alternate between merging Russian-into-Math and Math-into-Russian.

During the expert merging or expert upcycling training phase, we train on 2B tokens of Russian mathematics data extracted from web data using a combination of language identification (LID) and math classifiers. We found this additional cross capability in-domain training data was essential. Without it, all merged models struggle to achieve good cross capability performance (see experiments in \autoref{appendix:cross_capabilities}).

We introduce an additional baseline via continued pretraining the strongest dense model, the seed model, in a data-matched manner on the Russian mathematics data. This is to evaluate the impact of training on in-domain data without increasing the overall model capacity. Additional details of the experimental set-up are provided in \autoref{appendix:cross_capabilities}. All models are evaluated on the Russian subset of MGSM (8-shot; \citealt{shi2022language}), which are Russian translations of examples from GSM8K \citep{cobbe2021training}. 

\subsubsection{Cross capabilities results}

See \autoref{table:cross_capabilities} for cross capability results on Russian MGSM. Notably, we see that \BTS can effectively leverage both experts to excel at a new task, surpassing the data-matched seed model baseline, even though the experts themselves remain unchanged: by adding connections between them, the resulting model exceeds the sum of its individual parts. Among all expert-merging baselines, \BTS achieves the best cross capability performance. BTX and BAM variants also show strong performance, outperforming \BTS, likely due to their significantly greater training capacity on in-domain data. 

\subsection{\BTS architecture design}\label{sec:arch_design}

We ablate the impact of the number of stitch layers, the alternating stitch layer architecture, and the hub model selection.

\subsubsection{Impact of the number of stitch layers}

 \begin{table}[H]
\centering
\begin{tabular}{lrrrrrrrrr}
\toprule
& \multicolumn{2}{c}{\textbf{General}} & \multicolumn{2}{c}{\textbf{Code}} & \multicolumn{2}{c}{\textbf{Multilingual}} & \multicolumn{2}{c}{\textbf{Math}} & \\
\cmidrule(lr){2-3} \cmidrule(lr){4-5} \cmidrule(lr){6-7} \cmidrule(lr){8-9}
 & \textbf{MMLU} &
 \textbf{BBH} & \textbf{MBPP} & \textbf{HumanEval} & \textbf{Flores(S)} & \textbf{Flores(T)}  & \textbf{GSM8K} & \textbf{MATH} & \textbf{Avg.} \\ 
\midrule
10 Layers & \textbf{36.1} & 37.8 & 31.8 & 22.0 & 31.2 & \textbf{36.5} & 19.1 & 10.4 & 28.1\\
4 Layers & 35.8 & 36.9 & \textbf{32.2} & \textbf{22.0} & \textbf{33.9} & 36.2 & \textbf{20.2} & \textbf{10.6} & \textbf{28.1}\\
1 Layer & 34.9 & \textbf{37.8} & 29.6 & 19.5 & 30.8 & 35.9 & 17.7	& 9.9 & 27.0\\
\bottomrule
\end{tabular}
\caption{\textbf{Ablations on the effect of varying number of stitch layers on downstream task performance.} The first two rows are configurations with 10 and 4 stitch layers distributed uniformly throughout the seed and expert models. The third row is a configuration with a single Experts-into-Hub stitch layer placed after the last dense model layers. The 10 and 4 layers configuration performs similarly, but the single-layer configuration lags behind model performance significantly.}
\label{table:layers_ablations}
\end{table}

We measure the impact of varying the number of stitch layers on model performance, as shown in \autoref{table:layers_ablations}. The first two rows present configurations with $10$ and $4$ stitch layers, respectively, distributed uniformly throughout the seed and expert models.  In the third row, we investigate a configuration with a single Experts-into-Hub stitch layer placed after the final language model layers.

Our ablations show that a single stitch layer is insufficient for learning to effectively merge capabilities, as its performance lags significantly behind configurations with $4$ or $10$ layers. This also demonstrates that the \BTS models with more than one stitch layer combine models in a more expressive way than than simply combining output representations. The $4$ and $10$ layer configurations perform similarly, however, we note that this may be due to under-training of the $10$ layer variant as all models are trained on the same number of tokens.

\subsubsection{Importance of the alternating stitch layer architecture}

The \BTS architecture involves alternating between the Experts-into-Hub stitch layer and the Hub-into-Experts stitch layer.
We ablate the impact of adopting this alternating architecture, as opposed to utilizing all Experts-into-Hub layers. As shown in  \autoref{table:ablations_cross_capabilities}, the alternating architecture (first row) yields significantly better cross capability performance compared to using only homogeneous Experts-into-Hub stitch layers (second row). However, both the alternating and non-alternating architectures achieve comparable performance on generalist tasks, as shown in \autoref{table:layers_ablations2}. These results demonstrate that an alternating architecture is essential for achieving cross capability performance while maintaining strong generalist performance.

\begin{table}[h]
\centering
\begin{tabular}{lrrr>{\columncolor[gray]{0.9}}r}
\toprule
& & \multicolumn{2}{c}{\textbf{Flores}} \\
& \textbf{GSM8K} & \textbf{En/Ru} & \textbf{Ru/En} &  \textbf{Ru-MGSM} \\ 
\midrule 
\BTS Alternating & 13.3 & 31.9 & 34.7	& \textbf{16.0} \\
\BTS Experts-into-Hub Only & \textbf{15.2}& \textbf{32.0} & \textbf{35.0} & 11.6 \\
\bottomrule
\end{tabular}
\caption{\textbf{Comparison of alternating and non-alternating \BTS variants cross capabilities tasks} with additional in-domain Russian math training data. The alternating variant significantly outperforms the non-alternating variant.}
\label{table:ablations_cross_capabilities}
\end{table}

\begin{table}[htp]
\centering
\begin{tabular}{lrrrrrrrrrr}
\toprule
& \multicolumn{2}{c}{\textbf{General}} & \multicolumn{2}{c}{\textbf{Code}} & \multicolumn{2}{c}{\textbf{Multilingual}} & \multicolumn{2}{c}{\textbf{Math}} & \\
\cmidrule(lr){2-3} \cmidrule(lr){4-5} \cmidrule(lr){6-7} \cmidrule(lr){8-9}
 & \textbf{MMLU} &
 \textbf{BBH} & \textbf{MBPP} & \textbf{HE} & \textbf{Flores(S)} & \textbf{Flores(T)}  & \textbf{GSM8K} & \textbf{MATH} & \textbf{Avg.}\\ 
\midrule
\BTS Alternating & 35.8 & 36.9 & 32.2 & 22.0 & 30.9 & 36.2 & \textbf{20.2} & 10.6 & 28.1\\
All Experts-into-Hub & \textbf{36.1} & \textbf{37.9} & \textbf{32.4} & \textbf{22.6} & \textbf{31.4} & \textbf{36.4} & 19.9 & \textbf{10.8} & \textbf{28.4} \\
\bottomrule
\end{tabular}
\caption{\textbf{Comparison of alternating and non-alternating \BTS variants on generalist tasks.} Both variants achieves similar performance on most domains, with the non-alternating variant slightly outperforming the alternating variant on average. 
}
\label{table:layers_ablations2}
\end{table}

\subsubsection{Impact of hub model selection}

By default, we always use the seed model as the hub model in \BTS. 
This design choice is motivated from the fundamental nature of the seed model: as all experts are initialized from the seed model, the seed model's representations are more closely aligned with the experts' than the experts' are with each other, which may allow for more effective merging of representations via the \BTS stitch layers.

To validate this hypothesis, we conduct an ablation study in which we use an expert model as the hub instead. Specifically, we select the Math expert for this experiment, as it has the best generalist performance among all expert models. The seed model then is used as one of the ``experts'' or spoke models in \BTS. 
As shown in \autoref{tab:hub_ablations}, the results indicate that across most downstream tasks, selecting the seed model as the hub significantly outperforms using an expert model as the hub, validating this design choice. 
\begin{table}[htp]
\centering
\begin{tabular}{lrrrrrrrrrr}
\toprule
& \multicolumn{2}{c}{\textbf{General}} & \multicolumn{2}{c}{\textbf{Code}} & \multicolumn{2}{c}{\textbf{Multilingual}} & \multicolumn{2}{c}{\textbf{Math}} & \\
\cmidrule(lr){2-3} \cmidrule(lr){4-5} \cmidrule(lr){6-7} \cmidrule(lr){8-9}
 & \textbf{MMLU} &
 \textbf{BBH} & \textbf{MBPP} & \textbf{HE} & \textbf{Flores(S)} & \textbf{Flores(T)}  & \textbf{GSM8K} & \textbf{MATH} & \textbf{Avg.}\\ 
\midrule
Seed Hub & \textbf{35.8} & 36.9 & \textbf{32.2} & \textbf{22.0} & \textbf{30.9} & \textbf{36.2} & \textbf{20.2} & \textbf{10.6} & \textbf{28.1}\\
Math Hub & 33.9 & \textbf{37.8} & 30.7 & 20.1 & 29.8 & 36.0 & 15.6 & 5.73 & 26.2\\
\bottomrule
\end{tabular}
\caption{\textbf{Comparison of utilizing the seed model as the hub versus an expert.} We ablate \BTS (row 1) with a variant where we instead use the Math expert model as the hub (row 2). Using the seed model as the hub significantly outperforms using an expert model as the hub across most downstream tasks. This confirms that using the seed model as the hub in \BTS is important for achieving strong generalist performance.}
\label{tab:hub_ablations}
\end{table}

\subsection{Interpreting the \BTS stitch layers
}\label{sec:stitch_interpretation}
The gate values of the Experts-into-Hub stitch layer determine the weight of each expert in the combined representation.
Intuitively, the higher the expert or seed model's gate values, the more important this model is for the task. We inspect these values to get insight into the model's decision-making progress on various tasks.

\subsubsection{Visualizing gate values on expert specialty tasks}

\begin{figure}[htp]
    \centering
    \begin{subfigure}[b]{0.24\textwidth}
        \includegraphics[width=\textwidth, trim={0.4cm 0.9cm 0.4cm 0.9cm}, clip]{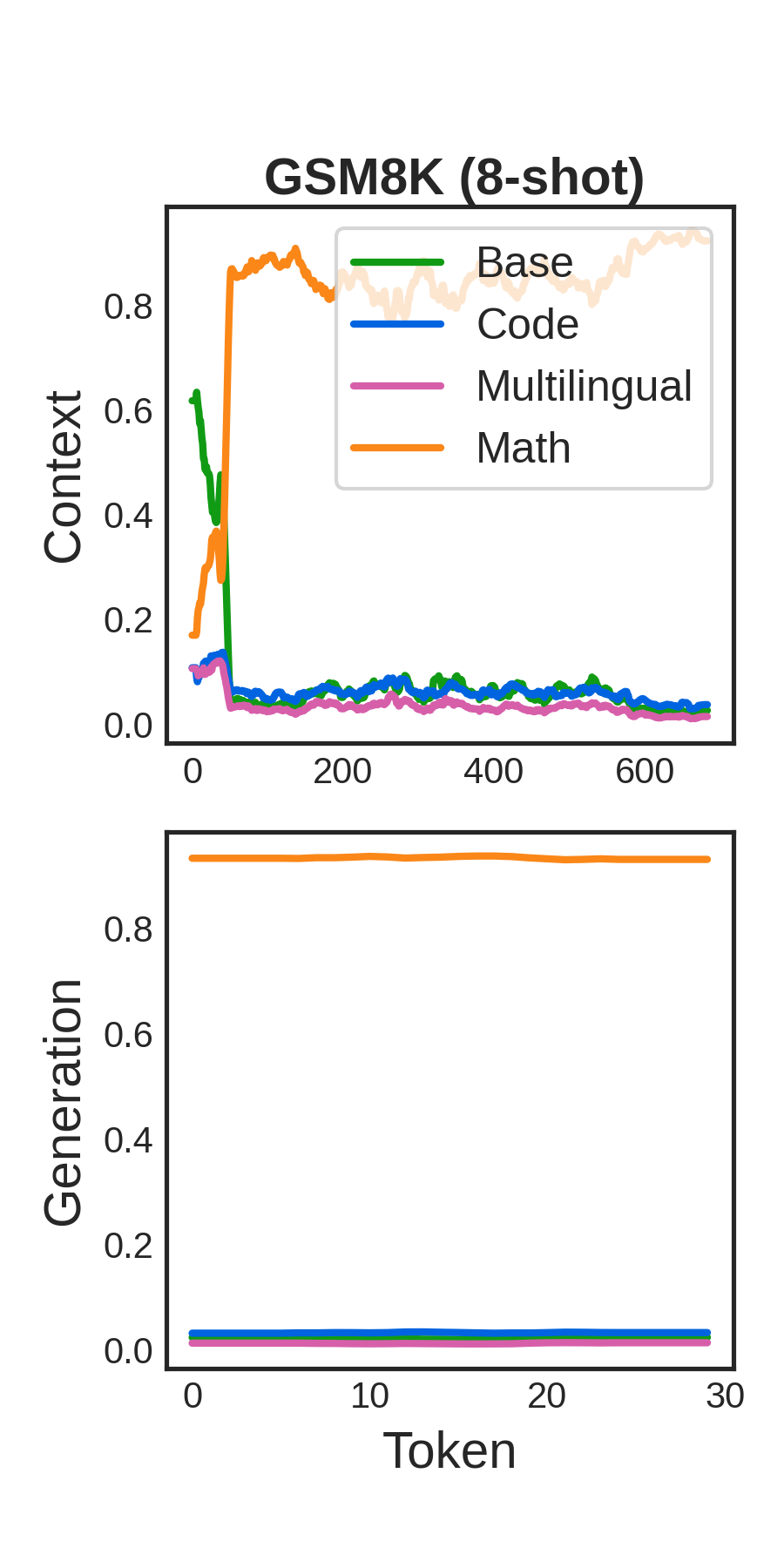}
        \label{fig:sub1}
    \end{subfigure}
    \hfill
    \begin{subfigure}[b]{0.24\textwidth}
    \includegraphics[width=\textwidth, trim={0.4cm 0.9cm 0.4cm 0.9cm}, clip]{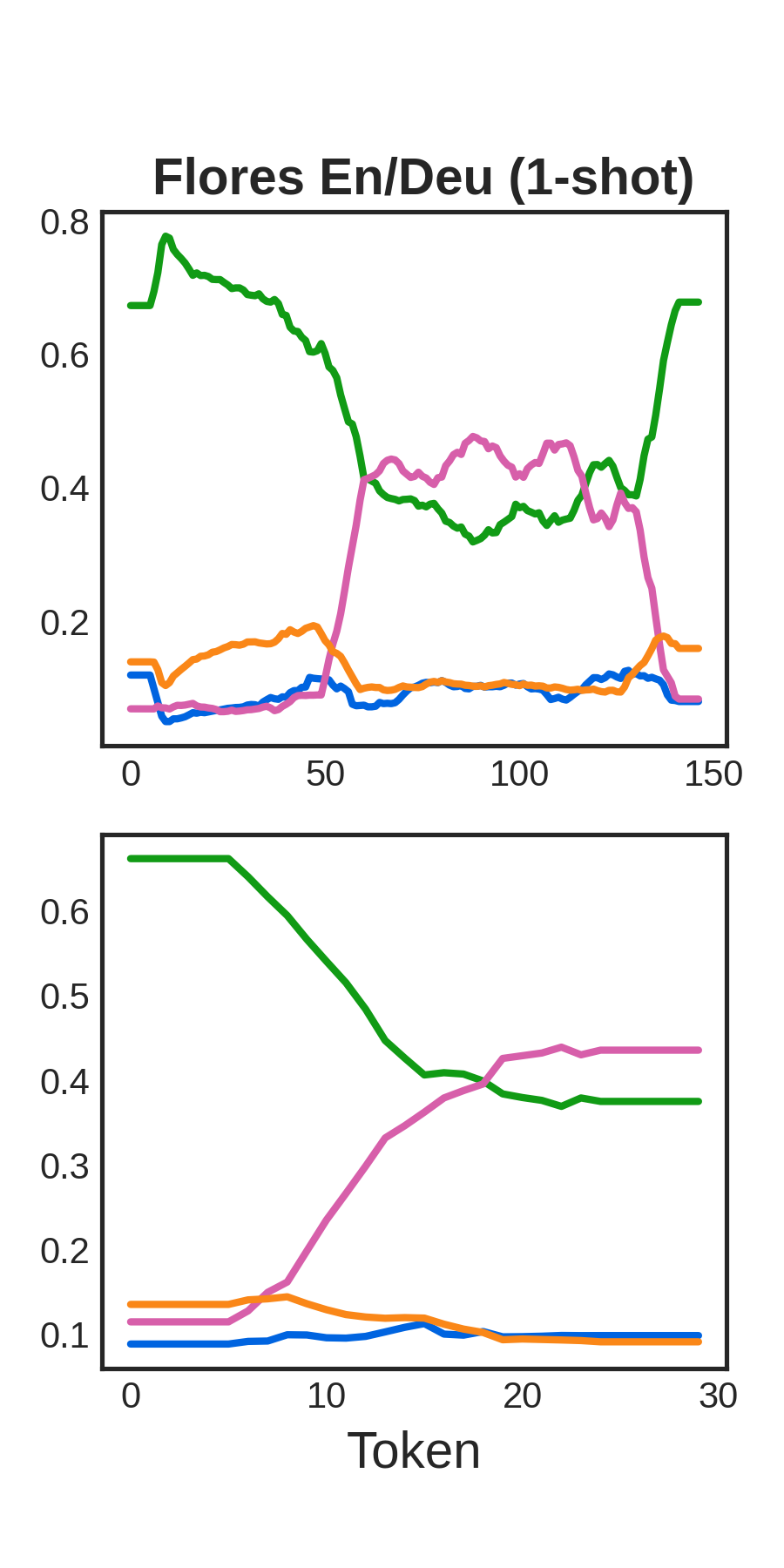}
        \label{fig:sub2}
    \end{subfigure}
    \hfill
    \begin{subfigure}[b]{0.24\textwidth}
    \includegraphics[width=\textwidth, trim={0.4cm 0.9cm 0.4cm 0.9cm}, clip]{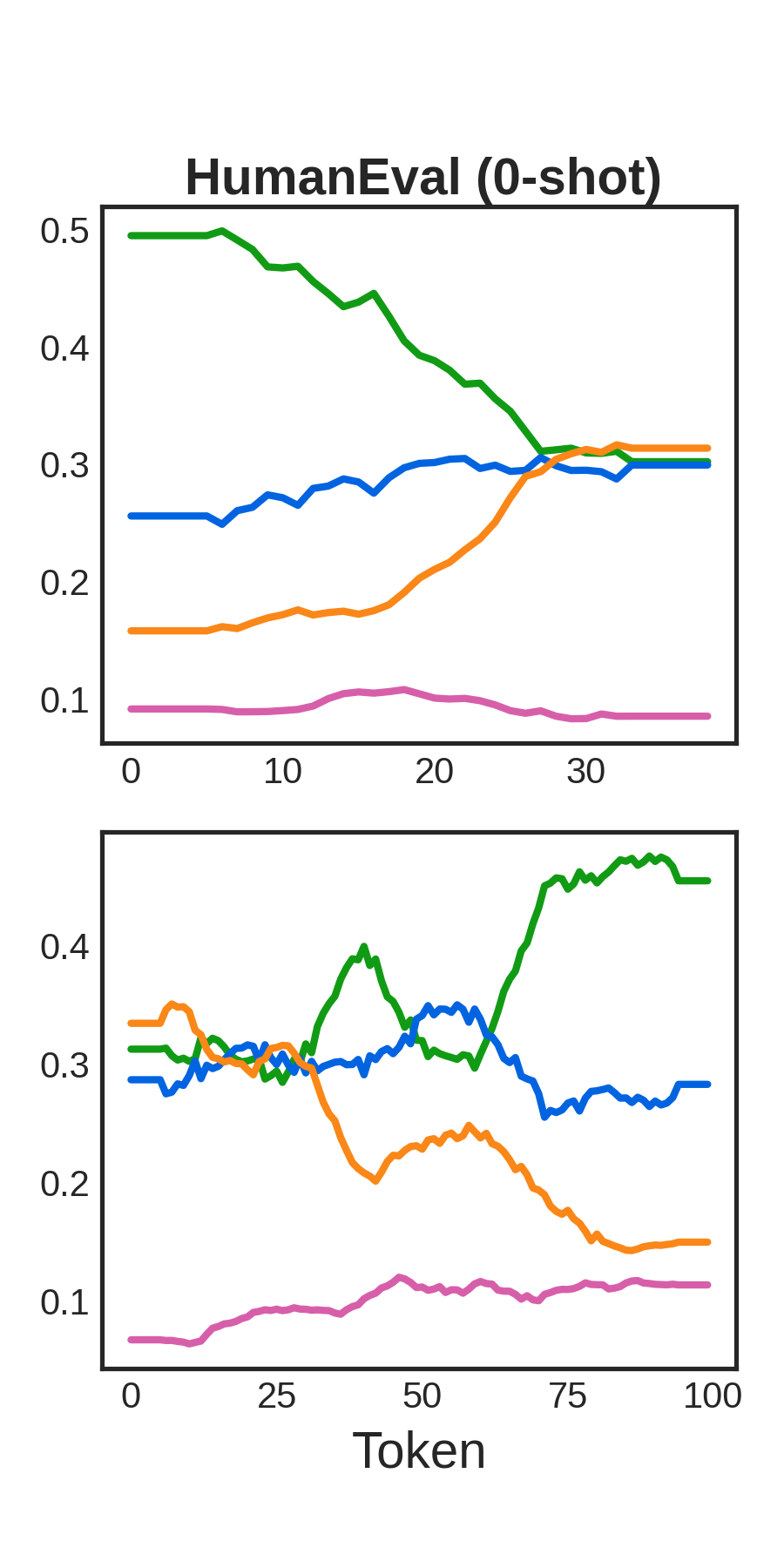}
        \label{fig:sub3}
    \end{subfigure}
    \hfill
    \begin{subfigure}[b]{0.24\textwidth}
    \includegraphics[width=\textwidth, trim={0.4cm 0.9cm 0.4cm 0.9cm}, clip]{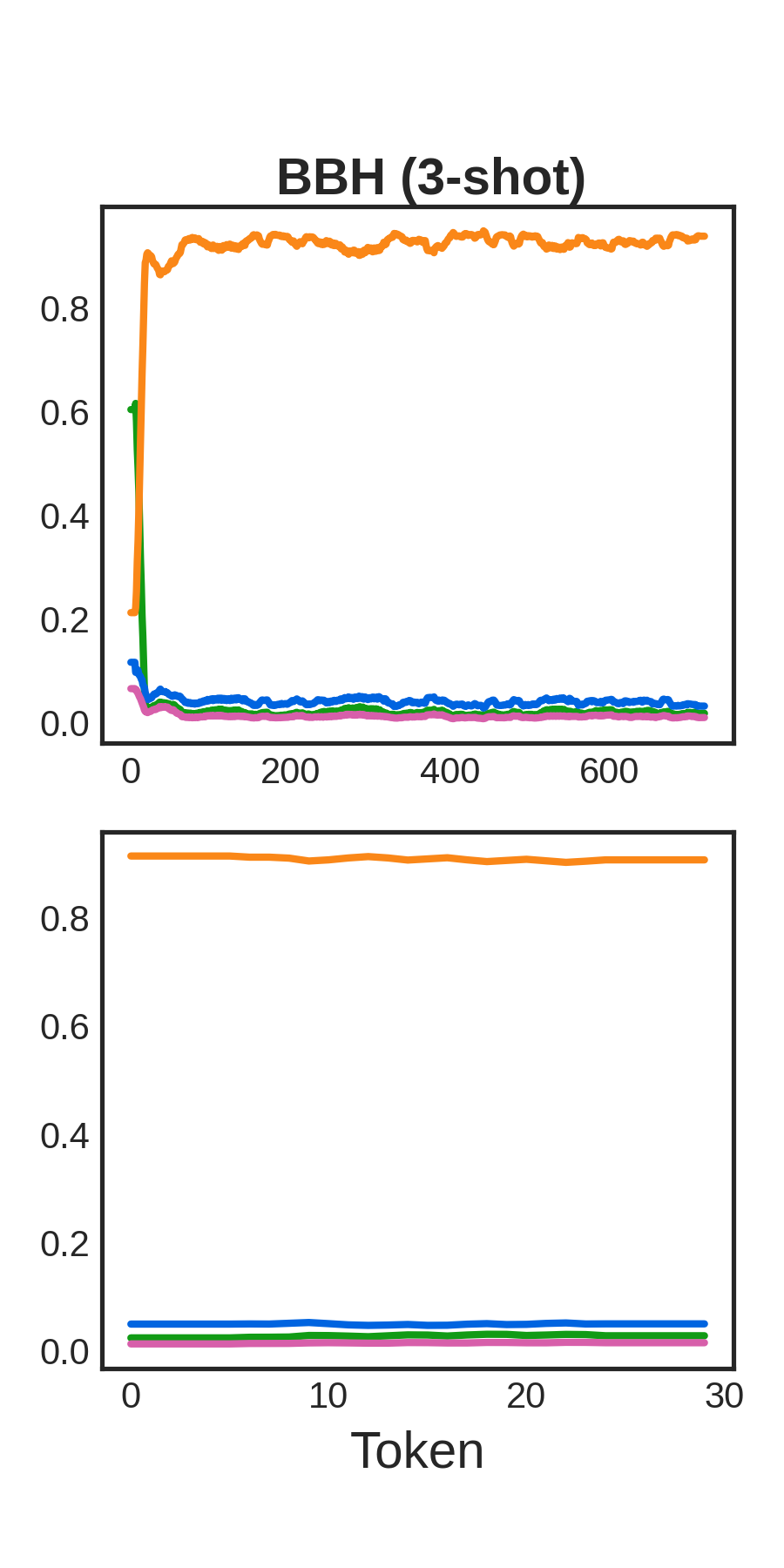}
        \label{fig:sub4}
    \end{subfigure}
    \caption{\textbf{Visualization of how \BTS gate values vary when generating a sequence during inference.} We inspect the gate values for the last stitch layer over the course of a sequence. The first row plots the gate values for prompt tokens, while the second row plots the gate values for the generated tokens. Each column corresponds to a different prompt, sampled randomly from the corresponding benchmark task. 
}
    \label{fig:gate_values}
    
\end{figure}

\autoref{fig:gate_values} visualizes how the gate values of the last stitch layer, an Experts-into-Hub stitch layer, vary when generating a sequence during inference on various expert specialty tasks. The first row plots the gate values for prompt tokens, while the second row plots the gate values for the generated tokens. Each column corresponds to a different prompt, sampled from the corresponding benchmark task.

This visualization shows that the gate values align closely with the task requirements -- with the specialized expert associated with the task typically dominating the gate values -- while effectively mixing representations from different models over the course of the sequence.  For example, for the the math task, GSM8K, the math expert has the highest gate value over the course of the generation while the other models' gate values are nearly zero. For language translation task, Flores, the multilingual expert and the seed model dominate, with each model being relied on more heavily at different parts of the prompt or generation.

\subsubsection{Visualizing gate value transitions on context-switching tasks}

\begin{figure}[ht]
    \centering
    \begin{subfigure}[b]{0.49\textwidth}
        \includegraphics[width=\textwidth, trim={0.3cm 0.5cm 0.3cm 0}, clip]{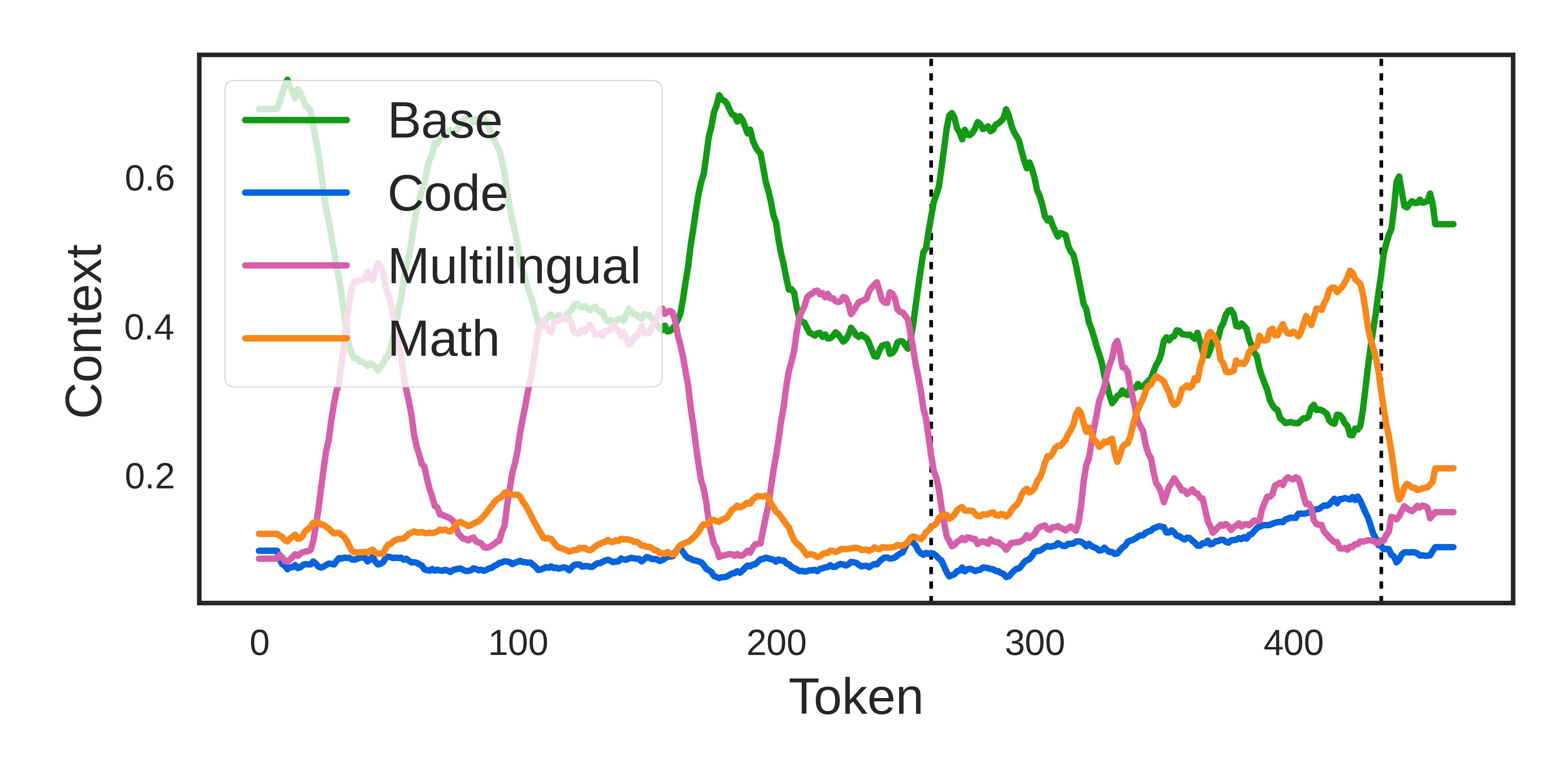}
        \label{fig:sub21}
    \end{subfigure}
    \begin{subfigure}[b]{0.49\textwidth}
    \includegraphics[width=\textwidth, trim={0.3cm 0.5cm 0.3cm 0}, clip]{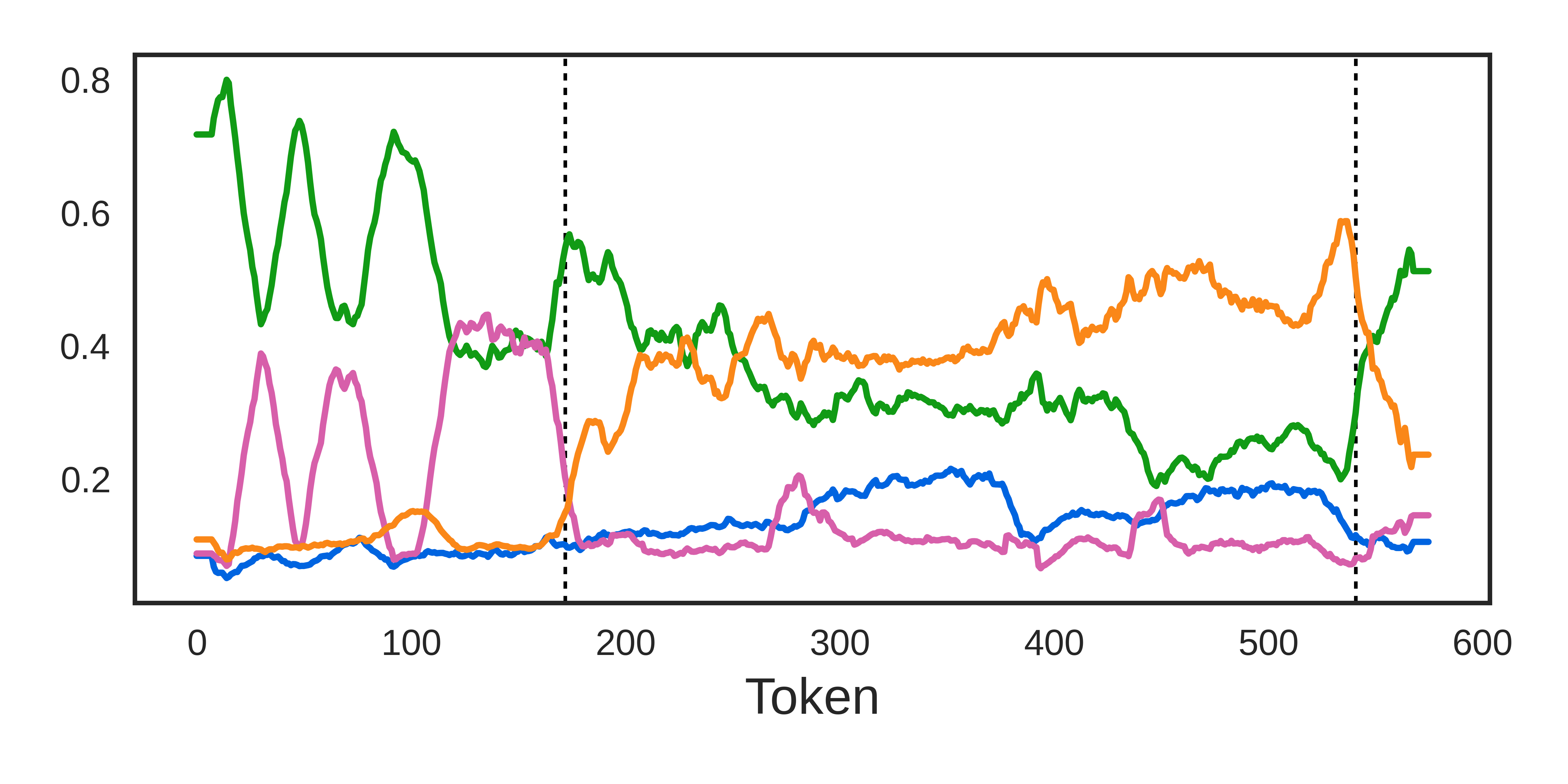}
        \label{fig:sub22}
    \end{subfigure}
    \caption{\textbf{Visualization of the gate values of \BTS's final stitch layer for context-switching sequences at inference time.} These sequences are constructed by concatenating question-answer examples from Flores (3-shot), GSM8K (2-shot), and TriviaQA (2-shot), in that order, with dotted lines indicating task transitions. Each plot corresponds to a different randomly sampled prompt. This visualization highlights \BTS's ability to dynamically adjust expert utilization based on token-level context.}
    \label{fig:gate_values2}
\end{figure}

Unlike merge methods which make sequence-level choices about which expert to use, \BTS can effectively context switch over the course of the sequence, seamlessly transitioning between different tasks. \autoref{fig:gate_values2} illustrates the gate values of \BTS's final stitch layer when processing context-switching prompts. These prompts are constructed by concatenating examples from Flores (3-shot), GSM8K (2-shot), and TriviaQA (2-shot) \citep{joshi2017triviaqa}, in that order, with dotted lines indicating where a new task begins. Each column corresponds to a different context-switching prompt, created from distinct sampled inputs.

In both examples, \BTS demonstrates its ability to dynamically adjust expert utilization. During the Flores prompt, the seed model and multilingual expert are predominantly active. During the GSM8K prompt, the math expert takes over, and finally, the seed model is most utilized for the TriviaQA prompt. This highlights \BTS's capability to correctly activate the relevant experts for each task, even when transitioning between diverse contexts.

\section{Related work}
\paragraph{Weights merging} 
Previous works have demonstrated that linearly interpolating the weights of multiple expert models with the same architecture can produce a more effective model. Model Souping \citep{wortsman2022model} achieves this by uniformly averaging model weights, whereas methods like BTM \citep{li2022branch}, C-BTM \citep{gururangan2023scaling}, and SMEAR \citep{muqeeth2023soft} dynamically compute the weighting of each expert's model parameters based on the given prompt. 

\paragraph{Output ensembles} 
In addition to averaging model weights, several works have explored averaging  model outputs to create ensembles of expert models \citep{li2022branch, gururangan2023scaling}. Unlike \BTS, these approaches do not require further training, and as such, they may be limited in expressivity.

\paragraph{Routing among dense models} 
Another approach involves routing the entire input and generation to a single model selected from multiple expert LLMs \citep{filippova2024no, ong2024routellm}. However, these methods are limited when the input requires expertise from multiple domains or involves context-switching between different tasks. In contrast, \BTS makes token-level decisions about combining experts, allowing it to seamlessly context switch across multiple tasks or adapt to inputs that require diverse or evolving skill sets.

\paragraph{Mixture-of-Experts upcycling} 

Several works have explored using pretrained dense models to initialize Mixture of Experts (MoEs) \citep{komatsuzaki2022sparse, sukhbaatar2024branch, zhang2024bam}. These approaches copy each expert model's parameters to initialize the corresponding experts in the MoE. For the MoE's non-expert parameters, they average the parameters of the pretrained experts. The router is initialized from scratch. Following initialization, the MoE undergoes a training phase where all model parameters are updated. On the other hand, \BTS only updates only the lightweight stitch layers, keeping all expert parameters frozen. Keeping experts intact after merging leads more more interpretable routing, and flexibility to add or remove experts with a small amount of additional training. 

\paragraph{Adding connections between models} Several recent works have proposed adapting language models to new modalities by composing modality-specific models, e.g., \cite{alayrac2022flamingo} propose adding cross-attention parameters to allow a language model to condition on visual inputs, and \cite{liang2024mixture} uses global self-attention to fuse models for different modalities. Perhaps most similar to our work, \cite{bansal2024llm} extend this idea to domain-specific language models, and propose augmenting an ``anchor'' language model with a single domain-specific model through cross-attention.  

\section{Conclusion}

We introduced \BTSlong, or \BTS, a simple, flexible method for merging expert models to create a stronger, unified, generalist model. \BTS combines expert models by inserting novel ``stitch'' layers between expert language model layers, which are learned in a lightweight training step. In experiments, we find that this approach outperforms competitive baselines, yielding the strongest generalist model performance with only a small number of training parameters. In some settings, \BTS is shown to even outperform the expert models in their specialized domains. We further demonstrate that a \BTS model can demonstrate new skills at the intersection of expert domains and motivate this architecture with extensive ablations and analysis. We hope this work furthers research into efficient and flexible methods for creating  generalist large language models from modular, independently-trained experts.

\section*{Acknowledgments}
We extend our thanks to Sachin Mehta, Ruslan Mavlyutov, Alexei Baevski, Onur Çelebi, Niladri Chatterji, and Mik Vyatskov for their assistance with the training infrastructure. We thank Vedanuj Goswami for his assistance in training the models used in the experiments. We are also grateful to Anirudh Goyal, Akhil Mathur, and Wenhan Xiong for providing helpful pointers regarding data mixtures. We appreciate Ellen Tan, Rocky Wang, Todor Mihaylov, Xuchao Jia, and Mihir Sanjay Kale for helping with using the data preparation pipeline. We also thank Florian Laplantif, Norman Cheng, Lovish Madaan, and Kunal Bhalla for their support with the evaluation infrastructure. QZ thanks Chris Lu for discussions on the project. Finally, we thank Melanie Kambadur for her general assistance and support of this project.

\clearpage
\newpage
\bibliographystyle{assets/plainnat}
\bibliography{main}

\clearpage
\newpage
\beginappendix

\section{Cross capabilities: additional details and experiments}\label{appendix:cross_capabilities}

\subsection{Further experiment details}
\paragraph{Russian expert model training}
To enhance cross capabilities in mathematical skills for Russian, we train an additional expert specifically on Russian data. The expert training setup follows the same procedure outlined in \autoref{sec:model_setup}. For training data, we utilize the Russian subset of the multilingual dataset previously used for the multilingual expert, as described in \autoref{sec:data_details}.

\paragraph{Merged models training}

As an additional baseline, we continually pretrain the strongest dense model, the seed model, on the same Russian math pretraining data used for the merged models. All experiments share the following training configuration, except for BTM and Model Soup which does not require further training:

\begin{itemize}
    \item Learning rate schedule: we warm up from $0$ to $5e-6$ over 1000 steps, then undergoes a cosine decay to 10\% of the peak learning rate. The merged models are trained for a total of 2000 steps. One exception is the expert routing model is trained for 1000 steps in total with a constant learning rate of 5e-4. This was chosen upon tuning the hyperparameters.
    \item Batch size: we use a batch size of 1M tokens. 
    \item Token count: All models were trained on 2B tokens of Russian mathematics over 2000 training steps. The exception is expert routing, which only trained on 1B tokens over 1000 steps, as we did not see performance improvement with further training.
    
\end{itemize}

\subsection{Results on merging and upcycling models without in-domain data}

\begin{table}[h]
\centering
\begin{tabular}{lrrr>{\columncolor[gray]{0.9}}r}
\toprule
& & \multicolumn{2}{c}{\textbf{Flores}} \\
& \textbf{GSM8K} & \textbf{En/Ru} & \textbf{Ru/En} &  \textbf{Ru-MGSM} \\ 
\midrule
\emph{\textbf{Dense models}} \\ 
Seed Model & 10.5	& 22.8& 32.8& \textbf{12.8}\\
Math Expert & $^*$\textbf{20.5}	& 10.2& 28.9& 10.8\\ 
Russian Expert &  9.48 &	$^*$\textbf{32.3} &	$^*$\textbf{34.6}	& 9.60\\
\midrule 
\multicolumn{3}{l}{\textbf{Expert upcycling}} \\ 
BTX Sample & 18.3 &	30.4 & 34.0	& 10.0 \\
BTX Soft & 18.0 &	30.0 & 33.9	& \textbf{12.4} \\
BAM & $^*$\textbf{20.5} & \textbf{30.6} & \textbf{34.5} & 10.8 \\
\midrule 
\multicolumn{3}{l}{\textbf{Expert merging}} \\ 
Model Soup & 17.5	&14.7	& 32.3	& $^*$\textbf{13.2}\\
BTM & $^*$\textbf{20.5} & $^*$\textbf{32.3}	& $^*$\textbf{34.6}	& 9.60 \\
Expert Routing & $^*$\textbf{20.5} &	\textbf{32.3} & $^*$\textbf{34.6} & 9.60 \\ 
BAM Adapter & 18.1 &	30.9 & 34.1	& 12.8\\
\BTS  & 19.0 &	31.6 & 33.0	& 10.0 \\
\bottomrule
\end{tabular}
\caption{\textbf{Cross capability performance of merged models \textbf{without} in-domain data.} We evaluate the seed model, Russian-language, and Math experts on Russian MGSM \citep{shi2022language} and compare performance with merged and upcycled models. We do not use any in-domain training data during the merging or upcycling training process. The results indicate that a small amount of cross capability data is necessary for merged or upcycled models to effectively learn cross capabilities.}
\label{table:cross_capabilities_without}
\end{table}

In \autoref{table:cross_capabilities_without}, we show results on merging and upcycling models without in-domain data. The merging phase is instead trained on a data mixture composed of 50\% of math expert and 50\% of Russian expert's continue pretraining data mixture. 

We observe that despite being trained with more tokens during the merging phase, all baseline methods does not significantly outperform the seed model on the cross capability task Russian MGSM. This indicates that in-distribution data is essential.

\section{Pseudo Code for \BTS}
\label{sec:pseudocode}

\begin{python}
def StitchLayer(xs, merge_into_hub=True):
    """
    xs: dense models' outputs
    """
    x_hub = x[0]
    x_experts = x[1:]

    g = w_gate(x_hub) # [bs, seq_len, dim, 1+n_experts]

    # Experts-into-Hub Layer
    if merge_into_hub:
        g = dropout(g).softmax(dim=-1)
        h_experts = [
            w_proj[i](x_experts[i]) for i in range(n_experts)
        ]
        h_hub = (g * stack([h] + h_experts, dim=-1)).sum(-1)

    # Merge-into-Expert Layer
    else:
        g = dropout(g).sigmoid()
        h_experts = = [
                (1 - g[..., i + 1]) * x_experts[i]
                + (g[..., i + 1] * w_proj[i](x_hub))
                for i in range(n_experts)
            ]
        h_hub = x_hub

    return stack([h_hub] + h_experts, dim=-1)

def BTSBlock(xs, ith_layer, BTS_freq):

    x_hub = hub_model_layer(xs[0])
    x_experts = [expert_model_layer[i](xs[i+1]) for i in n_experts]
    xs = stack([x_hub] + x_experts, dim=-1)
    
    if ith_layer 
        # Alternate between two types of stitch layers
        hs = StitchLayer(xs, merge_into_hub=(ith_layer//BTS_freq)
        return hs
        
    return xs

\end{python}

\end{document}